\documentclass[10pt,twocolumn,letterpaper]{article}

\usepackage[pagenumbers]{cvpr}      

\usepackage{listings}
\usepackage{algorithm}
\usepackage{algorithmic}
\usepackage{parskip}
\definecolor{cvprblue}{rgb}{0.21,0.49,0.74}
\usepackage{xcolor}
\usepackage[pagebackref,breaklinks,colorlinks,allcolors=cvprblue]{hyperref}

\newif\iffeedback
\feedbackfalse

\iffeedback
\newcommand{\sj}[1]{{{\textcolor{blue}{[SJ: #1]}}}}
\newcommand{\ba}[1]{{{\textcolor{red}{[BM: #1]}}}}
\newcommand{\nj}[1]{{{\textcolor{purple}{[NJ: #1]}}}}
\newcommand{\vb}[1]{{{\textcolor{brown}{[VB: #1]}}}}
\newcommand{\varun}[1]{{{\textcolor[HTML]{228B22}{[VC: #1]}}}}
\newcommand{\vv}[1]{{{\textcolor{teal}{[VV: #1]}}}}
\newcommand{\bn}[1]{{{\textcolor{magenta}{[BN: #1]}}}}
\else
\newcommand{\sj}[1]{}
\newcommand{\ba}[1]{}
\newcommand{\nj}[1]{}
\newcommand{\vb}[1]{}
\newcommand{\varun}[1]{}
\newcommand{\vv}[1]{}
\newcommand{\bn}[1]{}
\fi

\newcommand{\method}[0]{\textsc{MM-Gen}\xspace}

\DeclareMathOperator{\sref}{S_{T}^{ref}}
\DeclareMathOperator{\vpool}{V^{pool}_T}
\DeclareMathOperator{\types}{{types}_T}
\DeclareMathOperator{\srefk}{S_{T_k}^{ref}}
\DeclareMathOperator{\vpoolk}{V^{pool}_{T_k}}
\DeclareMathOperator{\ngen}{N_{gen}}

\def\paperID{10750} 
\def\confName{CVPR}
\def\confYear{2025}

\title{\resizebox{\textwidth}{!}{\method: Enhancing Task Performance Through Targeted Multimodal Data Curation}}

\author{
Siddharth Joshi$^{1,2}$\thanks{Work completed in part during an internship at Microsoft Research and in part as part of PhD thesis research at UCLA. Correspondence to {\tt\small sjoshi804@cs.ucla.edu}}, 
Besmira Nushi$^{1}$, Vidhisha Balachandran$^{1}$, \\
Varun Chandrasekaran$^{1,3}$, Vibhav Vineet$^{1}$, Neel Joshi$^{1}$, Baharan Mirzasoleiman$^{2}$ \\
$^1$Microsoft Research \quad $^2$UCLA \quad $^3$UIUC
}

\begin{document}
\maketitle
\newpage
\begin{abstract}
Vision-language models (VLMs) are highly effective but often underperform on specialized tasks, for example Llava-1.5 struggles on chart and diagram understanding, due to scarce task-specific training data. Existing training data, sourced from general-purpose datasets, fails to capture the nuanced details needed for these tasks. We introduce \method, a scalable method that generates task-specific, high-quality synthetic text for candidate images by leveraging stronger models. \method employs a three-stage targeted process: partitioning data into subgroups, generating targeted text based on task descriptions, and filtering out redundant and outlier data. Fine-tuning VLMs with data generated by \method leads to significant performance gains, including 29\% on spatial reasoning and 15\% on diagram understanding for Llava-1.5 (7B). Compared to human-curated caption data, \method achieves up to 1.6$\times$ better improvements for the original models, proving its effectiveness in enhancing task-specific VLM performance and bridging the gap between general-purpose datasets and specialized requirements. Code available at \url{https://github.com/sjoshi804/MM-Gen}. 
\end{abstract}
\vspace{-2mm}  
\section{Introduction}



\begin{figure*}[h]
    \centering
    \includegraphics[width=\textwidth]{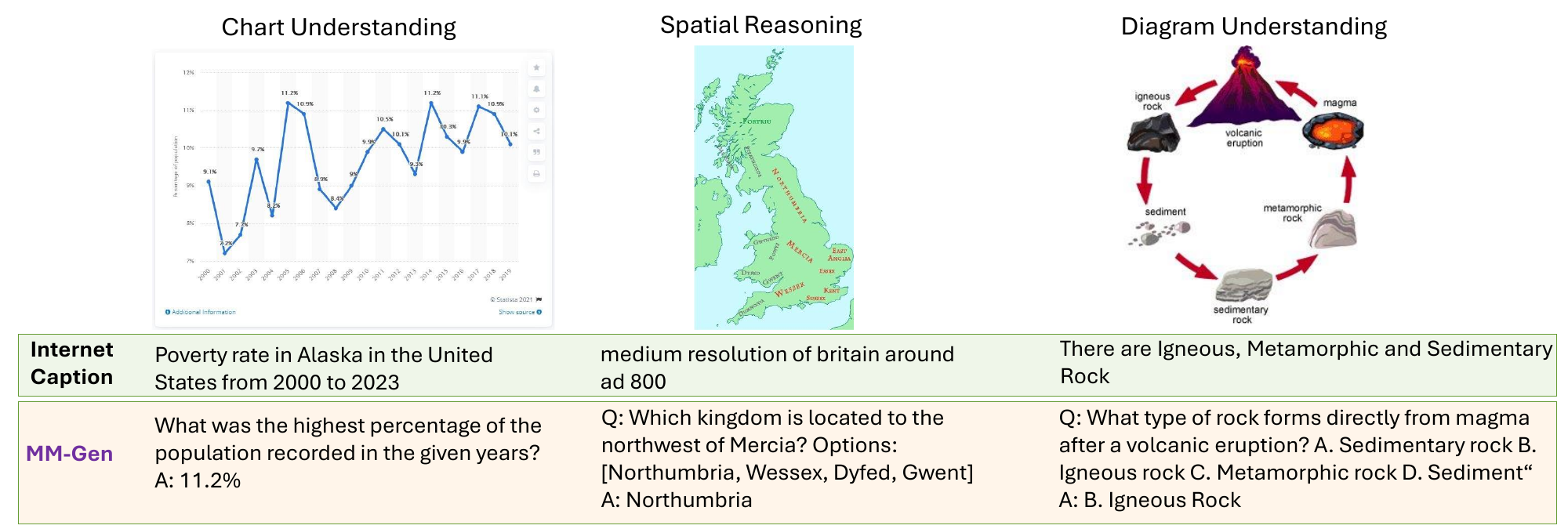}
    \caption{Examples of general text captions vs. task-specific text annotations generated by \method and used for fine-tuning supervision.}
    \label{fig:examples_annotations}
\end{figure*}

While vision-language models (VLMs) demonstrate state-of-the-art performance on several multi-modal tasks~\cite{llava}, they are often on tasks that are simpler in nature~\cite{balachandran2024eureka}. These models still struggle with more complex tasks, e.g., those that require fine-grained understanding of details in images~\cite{balachandran2024eureka, fu2024blink, kamath2023whatsupvisionlanguagemodels, stephens2024chatgpt}. 
We posit that the primary limitation for these VLMs is the quality and nature of the training data. VLMs are typically trained on large-scale image-text data scraped from the internet; while the images are rich and informative, the associated text descriptions can (i)  have limited relevance to the image~\cite{nguyen2024improving}, or (ii) omit references to several specific details captured in the image~\cite{lai2024revisitlargescaleimagecaptiondata}. Fig. \ref{fig:examples_annotations} shows examples of such images and web-scraped captions. While the images are relevant for the tasks of chart understanding, spatial reasoning, and diagram understanding, respectively, the text fails to capture details essential for these tasks. \looseness=-1

While synthetic caption generation strategies proposed in prior work~\cite{nguyen2024improving, lai2024revisitlargescaleimagecaptiondata, capsfusion} can create more descriptive text annotations (by referring to more visual details), they are agnostic of the downstream target task. Consequently, they {\em cannot} ensure that relevant details are captured in the text annotations. 
Recently, Shi et al.~\cite{mathllava} manually curated a task-specific dataset aimed at the task of multimodal mathematical question-answering by augmenting existing image-text data with detailed textual annotations, based on their domain expertise, using strong VLMs. While effective, such a curation pipeline involves significant human effort and is not scalable~\cite{chartinstruct, zhang2024multimodalselfinstructsyntheticabstract}. 

To address these limitations, we present \method{}, a highly general framework for \emph{automatically synthesizing task-relevant text annotations for images} by leveraging stronger VLMs (i.e., VLMs that perform well on the specific task) and requiring minimal human effort. \method{} takes as input a small set of examples from the target task (henceforth referred to as ``reference samples''), a list of image types associated with the task, and a pool of task-relevant candidate images for training. In practice, these inputs can be easily obtained: a small number of reference samples and associated image types can be directly collected from the target task, and a task-relevant image pool can be found via image search with search engines or retrieved from large-scale image-caption datasets \cite{schuhmann2022laion, sharma2018conceptual, changpinyo2021conceptual}. Using the reference samples to specify the details of the task to the stronger VLM, \method{} generates text-annotations that are task-relevant for the candidate images. Fig. \ref{fig:examples_annotations} shows how the text generated by \method captures task-relevant details. This simple approach is highly effective, resulting in significant improvements across a variety of target tasks. Moreover, human-effort in this process is limited to providing (i) a small set of reference samples for the task, and (ii) a pool of candidate images. To further improve the quality and efficacy of the generated data, \method{} 
introduces a perplexity \cite{brown1992introduction} based data-filtering approach to improve performance on target task using a high-value subset of the synthesized data. The components in \method{} are general and applicable to any image-text based target task enabling it to easily generalize across tasks and scale. 



We evaluate \method{} on improving VLMs' (e.g., Llava-1.5 7B and 13B parameter versions) performance on fine-grained image understanding tasks --- chart understanding and reasoning, diagram understanding, and spatial reasoning on maps. The data curated by \method{} enables an absolute improvement over Llava-1.5 (7B) of 15\%, 14\% and 29\%, respectively. We also see improvements over the much larger Llava-1.5 (13B) of 13\%, , respectively. Moreover, \method{}'s filtering strategy helps in shrinking data volumes by up to 50\% with no / minimal loss in performance. Empirical results show that models trained via \method{} data have a better performance than those trained via generated generic captions. \method data is also more effective than text annotations generated without task-specific reference examples, showing the importance of a targeted, data-centric approach for describing tasks. Finally, we analyze the effects of key design choices in \method through ablation studies on e.g. size of the reference sample set, generating with / without partitioning, scaling number of in-context samples.

\section{Related Work} 

\textbf{Synthetic Data Generation for Multimodal Models:} \citet{nguyen2024improving} highlighted the low quality of web-scraped captions. Later studies leveraged synthetic captions to enhance CLIP-style models \cite{fusecap, veclip, laclip, capsfusion}. Since these models are frequently applied to image classification, synthetic captions are usually general, task-agnostic descriptions that capture coarse-grained visual details about prominent objects in the image. We demonstrate in Sec. \ref{sec:experiments} that such data do not improve VLM performance in specialized tasks that require highly specific visual details.
Recent work has curated combinations of real and synthetic data to improve VLMs using stronger VLMs and hand-crafted prompts \cite{li2023m3it, sharegpt4v}. LLava \cite{llava} used manually designed prompts to generate diverse text annotations such as detailed descriptions, conversations, and complex reasoning. 
MiniGPT-4 \cite{minigpt4} first generates text using small VLMs and then
improves their quality by using strong VLMs, and finally manually filters the data. 
However, these data sets are not designed to improve performance on any specific task but instead focus on enhancing VLMs across a broad range of domains. More closely related is recent work on improving VLMs on specialized tasks. MathLLava \cite{mathllava} filters and augments human-curated multimodal data, using a stronger VLM and specialized prompts to create math VQA. Likewise, ChartInstruct \cite{chartinstruct} uses a highly specialized pipeline to generate chart-related data.
In contrast, our approach automates the generation of high-quality, task-specific text annotations, minimizing human intervention and generalizing across a  broader range of tasks. \looseness=-1

\textbf{Synthetic Data Generation for Training  LMs:}
\cite{tinystories, phi1, phi15, phi3, orca1, dubey2024llama} showed that LMs could be effectively pre-trained using high-quality synthetically generated data. 
Subsequently, \cite{orca2, orca3} highlighted the efficacy of synthetic task-specific training data to further improve performance in specialized tasks and proposed a framework to generate such data with minimal human intervention. We propose a generalizable framework for synthetic multimodal data generation to improve VLMs.

\textbf{Data Filtering Methods} Various filtering techniques have been explored for supervised learning \cite{coleman2019selection, toneva2018empirical, swayamdipta2020dataset, paul2021deep, katharopoulos2018not, mirzasoleiman2020coresets, pooladzandi2022adaptive, killamsetty2021grad}, self-supervised learning \cite{pmlr-v202-joshi23b, tripathidynamic}, and multimodal contrastive learning \cite{joshi2024dataefficientcontrastivelanguageimagepretraining, evans2024data, fang2023datafilteringnetworks, abbas2023semdedupdataefficientlearningwebscale, maini2024tmars}. More recently, data filtering has been applied to train generative LMs \cite{marion2023less, tirumala2023d, zhou2024lima, chen2023alpagasus, yang2023decoding}. We adapt filtering from \cite{marion2023less} to discard up to 50\% of data, with no / minimal drop in accuracy. \looseness=-1
\section{Problem Formulation}
\label{sec:prob}



Our objective is to generate text annotations for a given pool of candidate images, to improve performance, of a given VLM, on a target task $T$. 
Let a multimodal sample be denoted as $s = (v, t)$, where $v$ represents an image and $t$ represents the associated text (both the text prompt and text response). Let $\vpool$ denote the provided pool of candidate images, e.g., a corpus of chart images, and $\ngen$ the number of multimodal samples we wish to curate. Let $\sref$ be a small ($|\sref| = n \ll \ngen$) set of \textit{reference samples} that is \textit{representative} of the task $T$. This set serves as a reference for the text that is relevant for task $T$. In practice, this could be samples from the validation set of a dataset for chart understanding like ChartQA \cite{DBLP:conf/acl/MasryLTJH22}. Additionally, let $\types$ denote a list of the types of images associated with the task. For tasks like chart understanding, which have several different types of images, $\types$ could include \textit{bar charts}, \textit{line charts}, and \textit{pie charts}. The goal then is to use $\sref$, $\vpool$, and $\types$ to generate $\ngen$ multimodal samples for fine-tuning a given VLM, to improve performance on task $T$. To generate annotations, we assume access to a stronger VLM, i.e., one with higher performance than the given VLM on target task $T$ \footnote{In practice, this can be a VLM specialized on the task of interest (e.g., a VLM specialized for object detection if the task is detection), a general stronger model than the model of interest
or a combination of these.}. 

\section{\method Overview}
\label{sec_method}

In this section, we first motivate the need for task-relevant text annotations through an empirical case study. We then present \method{}: our framework for generating task-relevant text annotations.

\subsection{Case Study: MS COCO}
\label{sec:method:case_study}

Here, we illustrate using MS COCO~\cite{vinyals2016show} how {\em even human-crafted text annotations can omit important details in the accompanying image}. Recall that the MS COCO dataset is a high-quality and large-scale dataset commonly used for training models for image captioning \cite{santurkar2023is,nguyen2024improving}. Each image is manually annotated with 5 highly descriptive captions (annotators were asked to describe ``all relevant details''). 


\begin{figure}[!h]
    \includegraphics[width=0.9\linewidth]{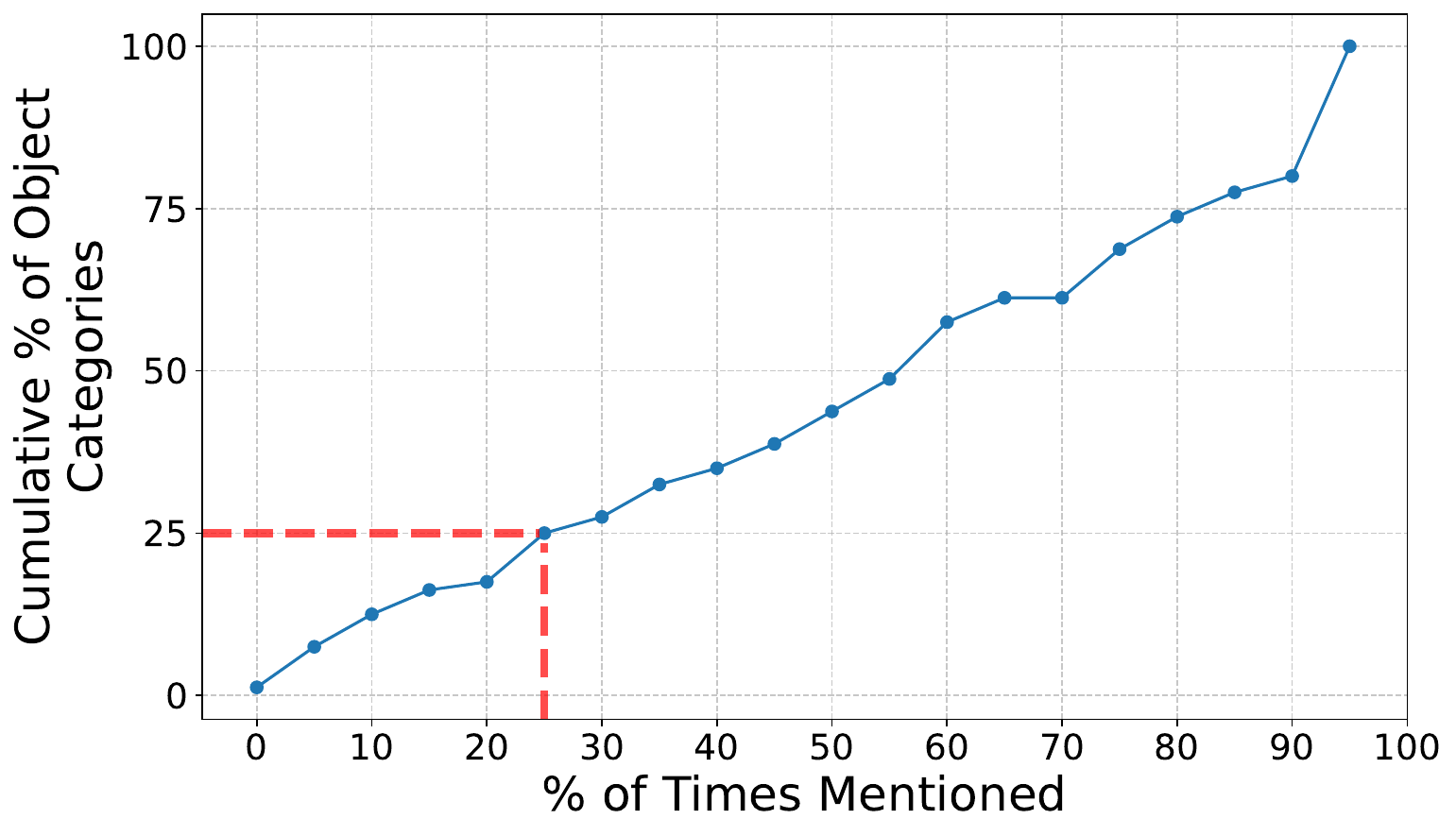}
    \centering
    \caption{Even high-quality human curated captions (MS COCO) miss many details found in images}
    \label{fig:mscoco}
\end{figure}\vspace{-2mm}

We now investigate how effective these captions are at capturing `all visual details' in the images. Fig.~\ref{fig:mscoco} shows the percentage of captions mentioning objects of different categories, when the corresponding image \emph{contains} the object. Remarkably, 25\% of object categories are included in the caption only for 25\% of the images (and are omitted 75\% of the time). For tasks that rely on recognition of objects from these categories, even MS COCO's ``high-quality'' captions will be ineffective.
This illustrates how even extremely descriptive, but task-agnostic text-annotations may be missing important information for tasks that require certain specific details about the image. Note that, the problem is different from the long-tail of visual concepts~\cite{changpinyo2021conceptual}; in this analysis, the content is present in the image but not in text. This observation motivates our claim that most general purpose text-annotations, human or VLM generated, are likely to omit specific details required by a particular downstream task. For example, descriptive captions of charts may still omit discussions on the minimum / maximum values, trends over time, values at specific points etc.

\subsection{\textbf{\method{}}: Design}



We now introduce \method{}: an automated way to generate text annotations that can provide task-specific supervision. Using the help of a running example i.e., curating multimodal data for \textit{Chart Understanding}, as exemplified by the ChartQA~\cite{DBLP:conf/acl/MasryLTJH22} dataset, we describe the 3 stages of \method{}: 1) partitioning data into subgroups, 2) generating task-relevant text annotations for each subgroup, 3) filtering the generated image-text pairs to keep the most informative ones. 


Chart understanding, exemplified by~\citet{DBLP:conf/acl/MasryLTJH22}, assesses a model’s reasoning skills on chart (in particular \textit{bar charts, pie charts and line charts}) visualizations using questions requiring logical, arithmetic, and data-driven interpretation. As a running example, we consider improving Llava-1.5-7B~\cite{llava} on task $T = \text{chart understanding}$, using a stronger VLM such as GPT-4~\cite{gpt4}. Here, the inputs to \method{} are:
\begin{enumerate}
    \item \textbf{Reference Sample Set} $\sref$: Examples from the ChartQA validation set. 
    \item \textbf{Types of Images} ${types}_T$ : \texttt{['bar chart', 'pie chart', 'line chart']}
    \item \textbf{Candidate Image Pool} $\vpool$: Corpus of chart images containing bar charts, pie charts and line charts.
\end{enumerate}

\subsubsection{Partitioning Data into Subgroups}\label{sec:method:partition}

Many tasks in multimodal learning are broad, covering a range of image types, each with unique text requirements within the same overarching ``task''. To capture this variability, we partition both the reference sample set and the candidate image pool into different subgroups, based on the various image types in this task, as delineated by $\types$ (e.g., \texttt{[`bar chart', `line chart', `pie chart']} for chart understanding). In practice, as we did, one can easily obtain $\types$ from a description of the dataset. 

\textbf{Why Partition?} Partitioning provides two key benefits. First, by grouping reference samples and candidate images by image type, we can identify, from the reference sample, the visual details that are task-relevant for a particular image type. In turn, we can effectively generate task-relevant annotations for candidate images of the same type. For example, on line charts, observing trends in y-axis values over a sequence of x-values is crucial, whereas on pie charts, the relative sizes of different slices are important. Thus, to generate task-relevant text for a line chart, using a reference of a line chart would be far more effective than using a reference from a pie chart. Second, we can ensure that the curated multimodal samples closely match the distribution of image types in the reference samples. Since the set of reference samples is representative of task $T$, matching the distribution of reference samples enables us to align the curated data with the distribution of task $T$.


\textbf{How to Partition?} Since we need to partition images based on the list of text in $\types$, a natural choice is the multimodal contrastive model CLIP~\cite{radford2021learningtransferablevisualmodels}. In particular, we leverage CLIP's zero-shot classification capabilities, as follows. Let $f_V$ and $f_T$ denote CLIP's vision and text encoders, respectively. We first encode the texts specified in $\types$ with the text encoder $f_T$. Next, we encode the images from $\sref$ and $\vpool$ using the vision encoder $f_V$. Then, we perform zero-shot classification on the images, matching each image to the text embedding with the highest cosine similarity ($S_C$) to its embedding:
$$k^* = \arg \max_{k \in \types} S_C\big(f_V(v), f_T(k) \big)$$
where $v$ refers to the image being classified (from the reference samples or candidate image pool), and $k$ refers to the $k$-th text in $\types$. This enables us to partition $\sref$ samples into subgroups based on the type of image they contain. $\sref = \{\cup_{k \in {\types}} \srefk\};\ \ \vpool = \{\cup_{k \in {\types}} \vpoolk\}$
where each $\srefk$ contains images of type $k$ (where $k \in types$) as well as the corresponding text annotations, and each $\vpoolk$ contains candidate images of type $k$. 







\begin{figure*}[t]
\includegraphics[width=\textwidth]{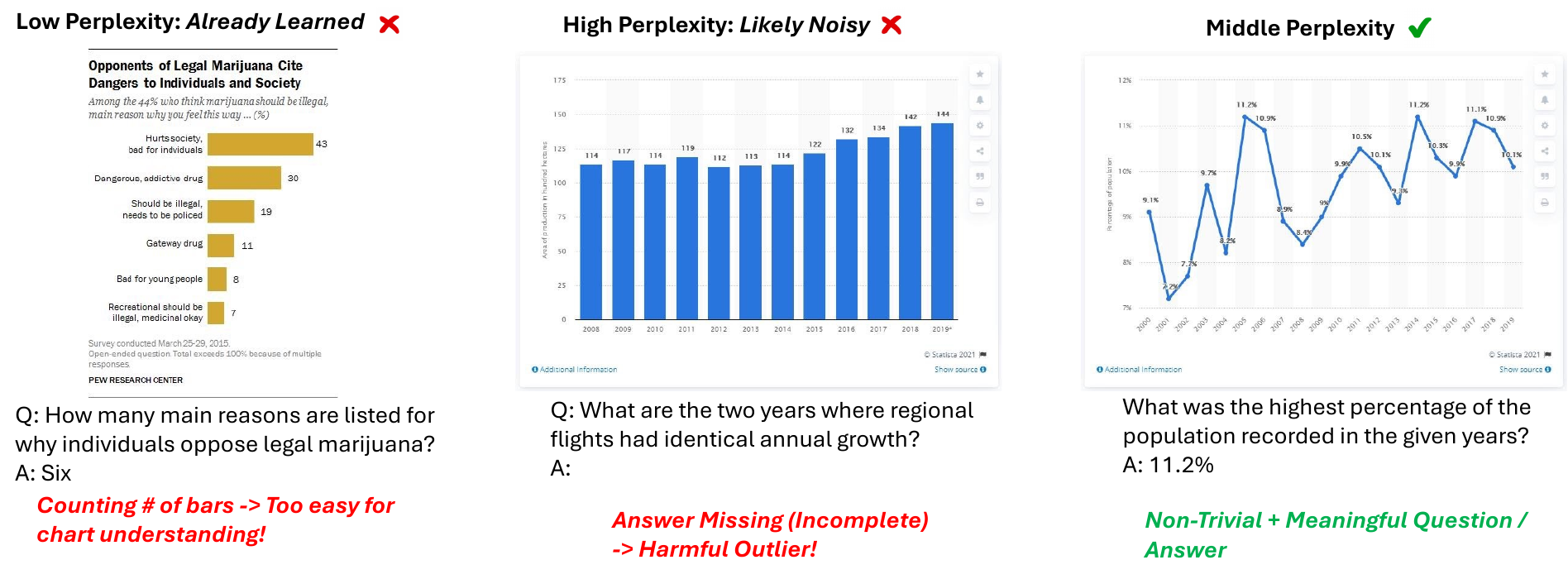}
\caption{Examples of different text perplexity mapping to easy cases (low perplexity), potential noise and outliers in difficulty (highest perplexity), and meaningful, non-trivial questions (middle perplexity). Questions with middle perplexity are also more likely to add new and useful training signal.}
\label{fig:examples_perplexity}
\end{figure*}\vspace{-2mm}

\subsubsection{Generating Task-Relevant Text Annotations}
\label{sec:method:response_generation}

Now we present how we generate task-relevant text annotations by leveraging a stronger VLM. 


Recall that in \S~\ref{sec:method:case_study}, we motivated the importance of task-specific text annotations. Here, we consider how to generate such annotations using a stronger VLM. In particular, we present how to describe the task $T$ to the stronger VLM. An obvious solution is using a natural language description of the task. However, this requires significant human effort to accurately and comprehensively describe the task. This prevents efficiently generalizing to new tasks. Moreover, even with significant human effort, describing the task in such a manner is extremely challenging and can easily lead to an under-specified task description. For example, consider a natural description of chart understanding as ``the ability to understand and reason over chart-based visualizations, focusing on bar charts, line charts and pie charts'' (this description is created by~\citet{DBLP:conf/acl/MasryLTJH22}). However, such a description does not capture the visual details required to answer detailed questions on charts (e.g., attention to visual depictions of trends could be task-relevant for line charts). 

A second, far more reliable and effective approach is \textit{data-centric} i.e., use the reference samples to specify the task to the stronger VLM. In particular, we can leverage the in-context learning ability of stronger VLMs to use reference samples from task $T$ to indicate to the model what types text is relevant for task $T$. In-context learning refers to large generative models' ability to learn tasks from a minimal number of examples or demonstrations, often with just one~\cite{dong2024surveyincontextlearning}. This approach 1) signficantly reduces the requirement on human effort to selecting a small \# of reference samples e.g., for chart understanding, these can be examples from the validation set of ChartQA, and 2) ensures the task is sufficiently specified since reference samples demonstrate which visual details are task-relevant. In \S~\ref{sec:experiments}, we conclusively demonstrate the superiority of the second approach for describing the task to the stronger VLM.

For each subgroup $(\srefk, \vpoolk)$, we generate text annotations by randomly sampling a reference sample for the subgroup and a candidate image for the input to the stronger VLM. The exact construction of the prompt and resulting examples are included in Appendix~\ref{app:examples} and Figure~\ref{fig:examples_annotations}. Given $\ngen$, the target size for the generated data, for each subgroup in the partition  $(\srefk, \vpoolk)$ where $k \in \types$, we generate a fraction of $\ngen$, proportional to the size of the reference samples of the subgroup. We then pair the candidate images, with the corresponding generated text-annotations i.e., the text prompts and the text responses, to curate the desired $\ngen$ multimodal samples. 

\subsubsection{Filtering Generated Data} \label{sec:method:filtering}


We now present two important considerations that motivate filtering the curated data before training the VLM. First, the VLM we wish to improve may have non-trivial initial performance on task $T$ and hence, it is possible that it might already fit some of the generated data effectively, i.e., generate accurate responses given the image and text prompt without even training on them. In this case, training on these data is unlikely to provide a significant improvement on the target task $T$. Second, since the stronger VLM is not necessarily perfectly accurate on task $T$, some of the generated examples may be unhelpful outliers (may have incorrect or incomplete responses, malformed questions etc.). Training on such data will not improve the performance, but may also degrade the performance of the VLM we wish to improve. 

To filter out such examples from the generated data, we rely on a filtering criterion from data-filtering for LLM pre-training~\cite{marion2023less} that captures a similar notion: selecting examples with \textit{middle perplexity}. Perplexity is a measure of how well a probability model predicts a sample
in language modeling tasks~\cite{brown1992introduction}. Formally, given a sequence of tokens \( w_1, w_2, \dots, w_n \) with probability \( P(w_1, w_2, \dots, w_n) \), the perplexity is defined as:
$\exp\left(-\frac{1}{n} \sum_{i=1}^n \log P(w_i | w_1, \dots, w_{i-1})\right)$.
A lower perplexity indicates that the model is already fit well to the tokens, while a higher perplexity implies greater difficulty in fitting the tokens~\cite{jelinek1997statistical}. Thus, by definition, samples that have been already fit have low perplexity. Moreover, outliers will have higher perplexity; by discarding outliers we can minimize the number of useless / harmful (incomplete / incorrect / malformed) examples in the generated data . In our framework, we measure the perplexity of each instance using the masked language modeling (MLM) objective, computing this over the text response, conditioned on the image and text prompt as input, using the VLM we wish to improve. Fig. \ref{fig:examples_perplexity} shows examples of low, medium and high perplexity samples.


Empirically, we retain 50\% of the generated data through filtering. This demonstrates significant gains in both performance and efficiency across tasks. Extended experimentation on optimizing the best fraction of data to retain per task may provide further benefits, albeit here we keep this fraction constant for all tasks for simplicity. 

\section{Experiments}\label{sec:experiments}\vspace{-1mm}


\textbf{Tasks.} We evaluate \method{} on 3 complex multimodal tasks, requiring fine-grained understanding of details in the images, that several existing VLMs struggle on: 1) chart understanding \& reasoning, 2) diagram understanding, and 3) spatial reasoning on maps.

\textit{Chart Understanding and Reasoning:} We use ChartQA~\cite{DBLP:conf/acl/MasryLTJH22} to evaluate the ability of a model to understand and reason over chart-based visualizations. As inputs to \method{}, we have: 1) Reference Samples: the validation set of ChartQA ($\approx$ 1K samples); 2) Types of Image: determined from dataset description as \texttt{[`bar chart', `line chart', `pie chart']}; 3) Candidate Image Pool: 15K images of charts taken from the ChartQA training set. With these inputs, we curate 150K multimodal samples and retain 75K after filtering. 


\textit{Diagram Understanding:} We use AI2D Diagrams (AI2D)~\cite{kembhavi2016diagram} to asses a model’s diagrammatic understanding using grade-school science diagrams and associated multiple-choice questions about the relationships and components in these diagrams. As inputs to \method{}, we have: 1) Reference Samples: a random subset of size 100 sampled from AI2D's training set; 2) Types of Image: determined as \texttt{[`physics diagram', `biology diagram', `chemistry diagram', `geography diagram']} from the dataset description; 3) Candidate Image Pool: approximately 5K diagram images taken from the training images of AI2D. With these inputs, we curate a total of 100K multimodal samples and retain 50K after filtering.

\textit{Spatial Reasoning on Maps:} We use SpatialMap~\cite{wang2024picture} to test the spatial reasoning capabilities of VLMs on maps by requiring them to answer questions on cardinal directions (e.g., North, South, East, West) and reasoning about the relationships between different landmarks in the map. As inputs to \method{}, we have: 1) Reference Samples: the validation set of SpatialMap; 2) Types of Image: determined from dataset description as \texttt{[`map']}; 3) Candidate Image Pool: 1K images of maps retrieved from DataComp-Small~\cite{gadre2024datacomp} using CLIP embedding search. With these inputs, we curate 50K multimodal samples and retain 25K after filtering.

\begin{figure*}[t]
    \includegraphics[width=\textwidth]{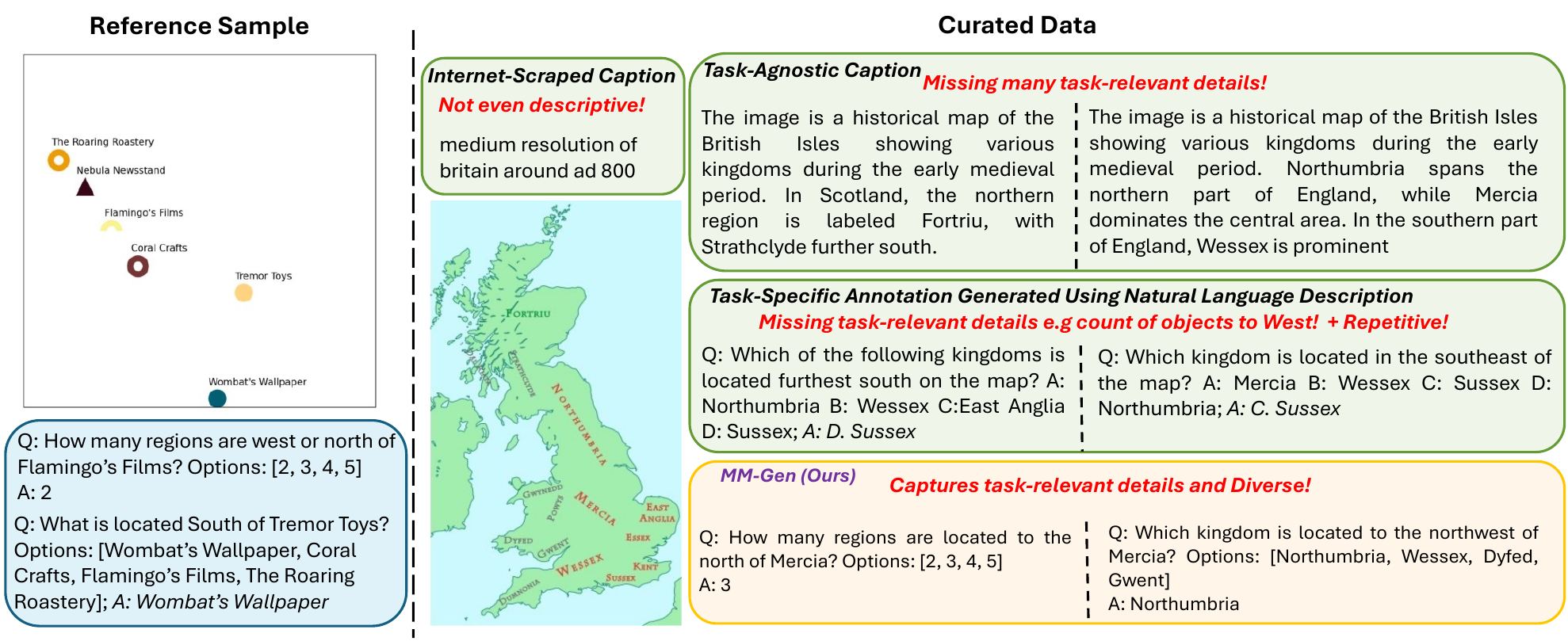}
    \caption{Comparing different baselines for multimodal data generation with \method{}. \method{} not only customizes the generated text to the task via reference samples, but it also adds missing details to the text that are required for answering the task.}\label{fig:compare_baselines}
\end{figure*}

\textbf{Baselines and Skyline.} Since \method{} is the first framework for curating task-specific multimodal samples, we contribute baselines and a skyline to evaluate its effectiveness. We use GPT-4o~\cite{gpt4} as the stronger VLM to generate the text annotations. Exact inputs and generated examples appear in App. \ref{app:examples}. We enumerate them below:

\textit{1. Base Model}: This refers to the initial performance of the VLM, before any additional training. 

\textit{2. Task-Agnostic Captions}: This baseline uses task-agnostic text annotations generated by a stronger VLM for the candidate image pool. This is to compare with how traditional caption generation methods, that do not generate text annotations specialized for tasks, would perform on our tasks. 

\textit{3. Task-Specific Text Annotations, Generated using Natural Language Task Description but no Reference Images}: This baseline uses text annotations generated by a stronger VLM to be task-specific using a natural language description of each task. These descriptions are obtained from the original dataset descriptions~\cite{DBLP:conf/acl/MasryLTJH22, kembhavi2016diagram, wang2024picture}. This comparison allows us to compare the effectiveness of describing the task using natural language descriptions vs. describing using reference samples from the task, as is done by \method{}.

\textit{4. Skyline -- Training on i.i.d. Training Data}: When i.i.d. training data (curated manually by humans) specifically for the target task includes task-relevant details, training directly on this data creates as a skyline model; this provides a performance benchmark for \method{} to approach or surpass. For 1) ChartQA, the skyline is i.i.d training data of size $\sim 30K$ containing images of charts, coupled with chart understanding question-answers; 2) AI2D, the skyline is i.i.d training data of size $\sim 5K$ containing images of grade-school diagrams, coupled with diagram understanding questions; 3) SpatialMap, the skyline is generated using the code provided to generate the evaluation set, containing images of synthetically generated maps, coupled with spatial reasoning question-answers. 

\textbf{Models.} As the target VLM to improve, we use Llava-1.5 (7B parameters)~\cite{llava}, comparing the performance of the base model (before training on any additional data) to that of training on the data curated by the aforementioned baselines, the skyline and \method{}. To investigate the effectiveness of our approach across model sizes, we additionally evaluate \method{} on Llava-1.5 (13B parameters). Details on training setup are presented in Appendix \ref{app:hparams}.

\subsection{Analysis of Performance of across Tasks}

\begin{figure*}[h]
    \centering
    \begin{subfigure}[b]{0.33\textwidth}
        \centering
        \includegraphics[width=1.03\linewidth]{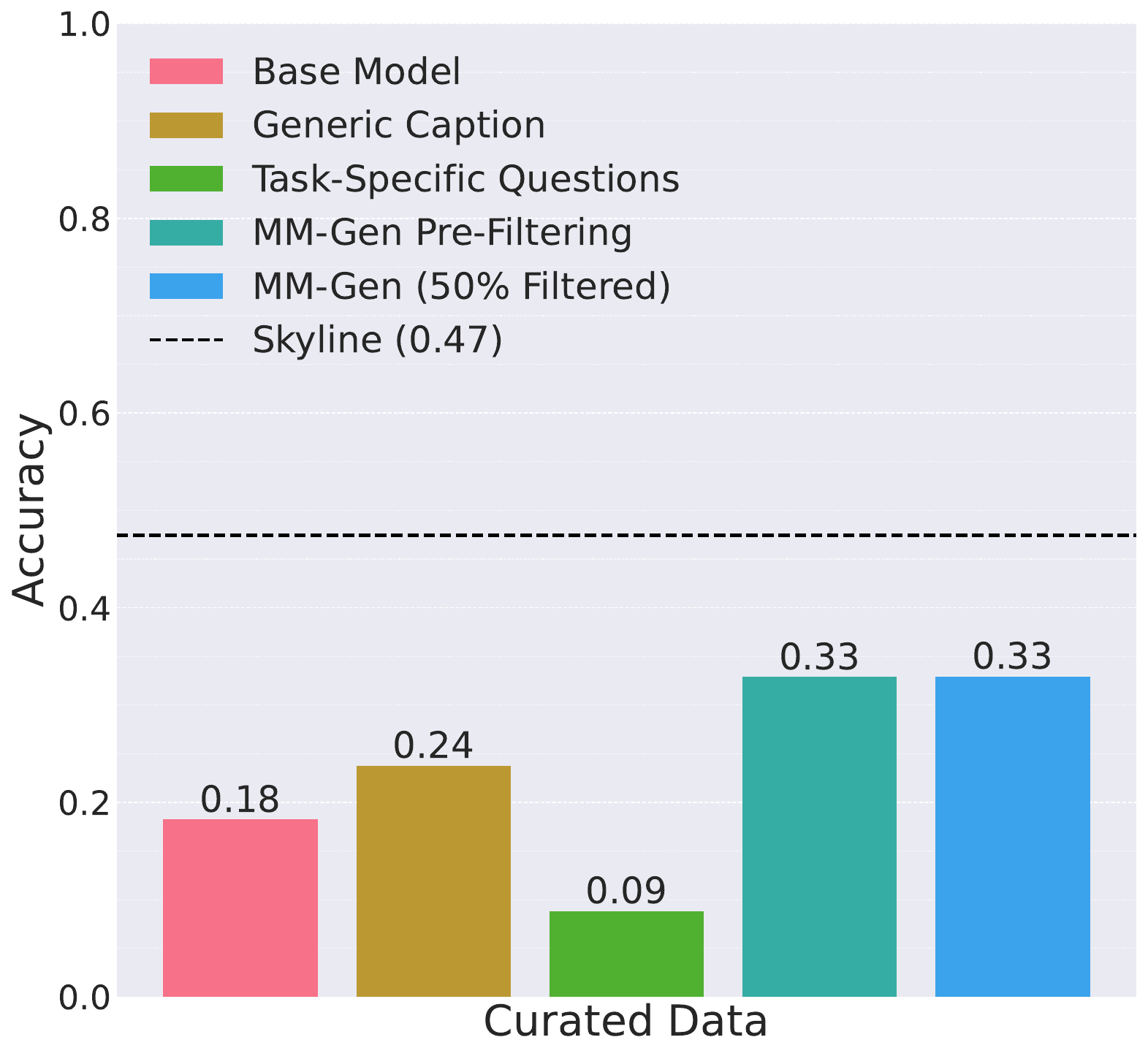}
        \caption{ChartQA}
        \label{fig:subfig1}
    \end{subfigure}
    \hfill
    \begin{subfigure}[b]{0.33\textwidth}
        \centering
        \includegraphics[width=1.03\linewidth]{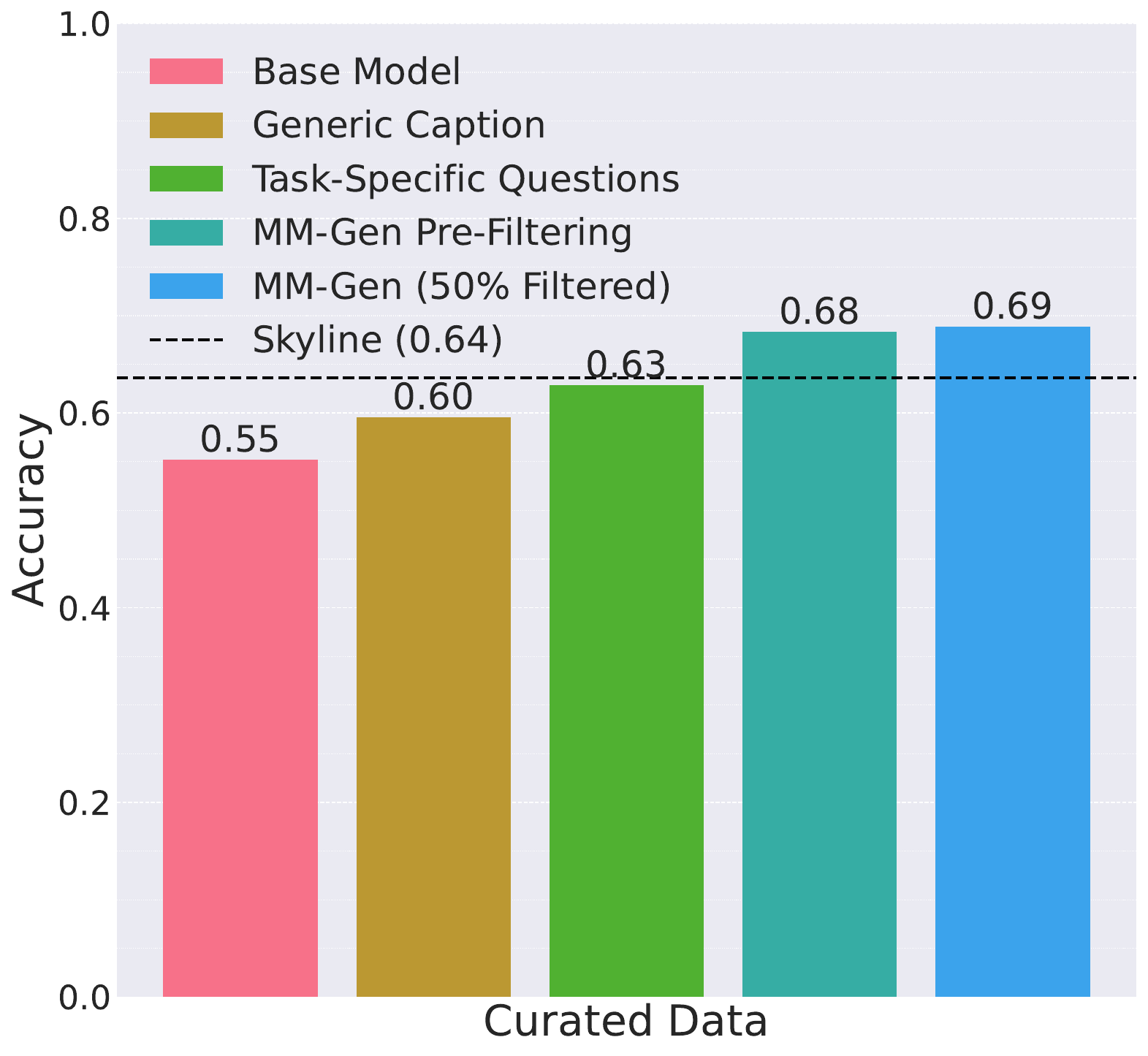}
        \caption{AI2D}
        \label{fig:subfig2}
    \end{subfigure}
    \hfill
    \begin{subfigure}[b]{0.33\textwidth}
        \centering
        \includegraphics[width=1.03\linewidth]{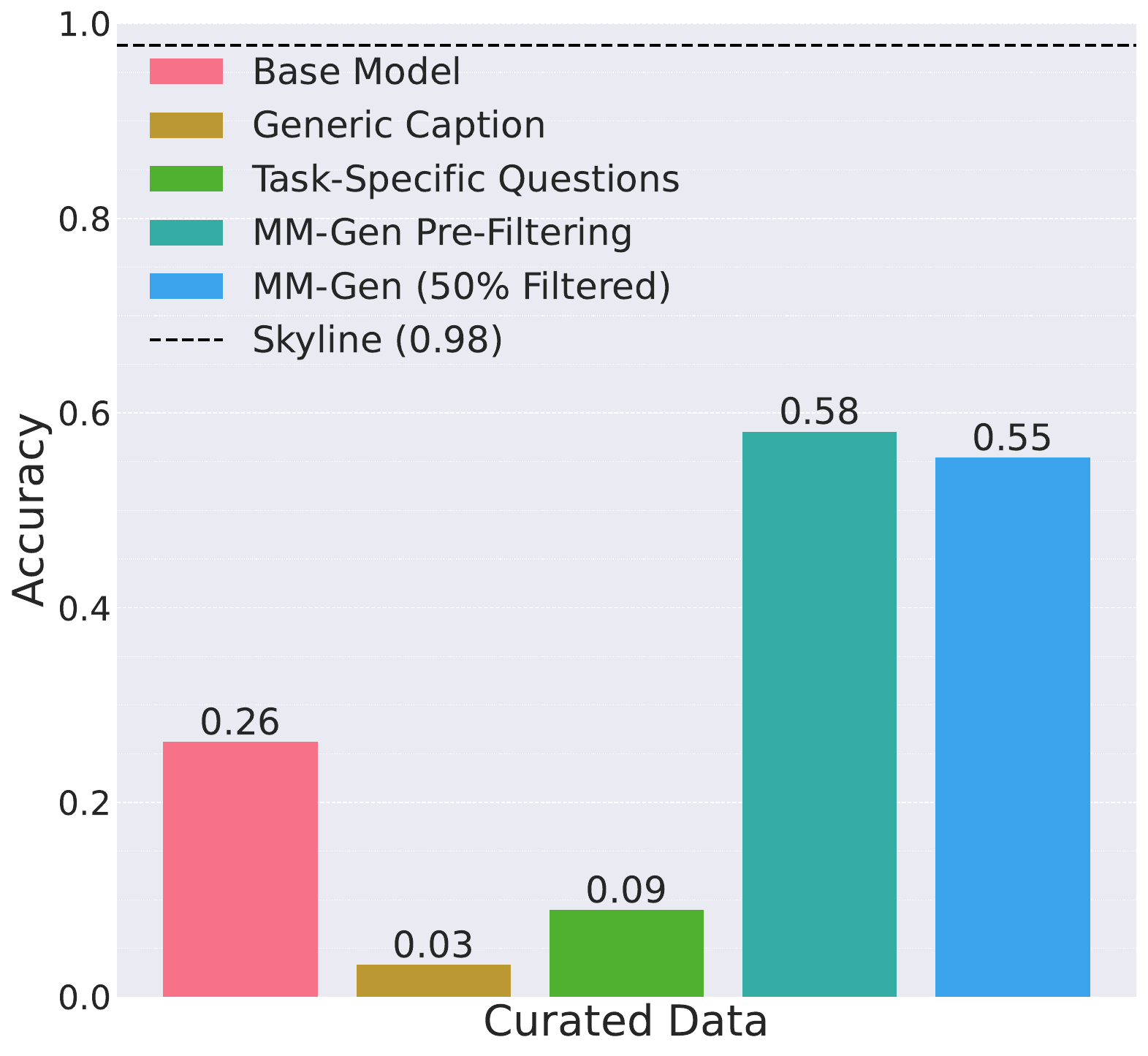}
        \caption{SpatialMap}
        \label{fig:subfig3}
    \end{subfigure}
    \caption{Comparing performance of \method\ across Tasks against Contributed Baselines and Skyline}\vspace{-2mm}
    \label{fig:main_results}
\end{figure*}

\begin{figure}
    \centering
    \includegraphics[width=0.95\linewidth]{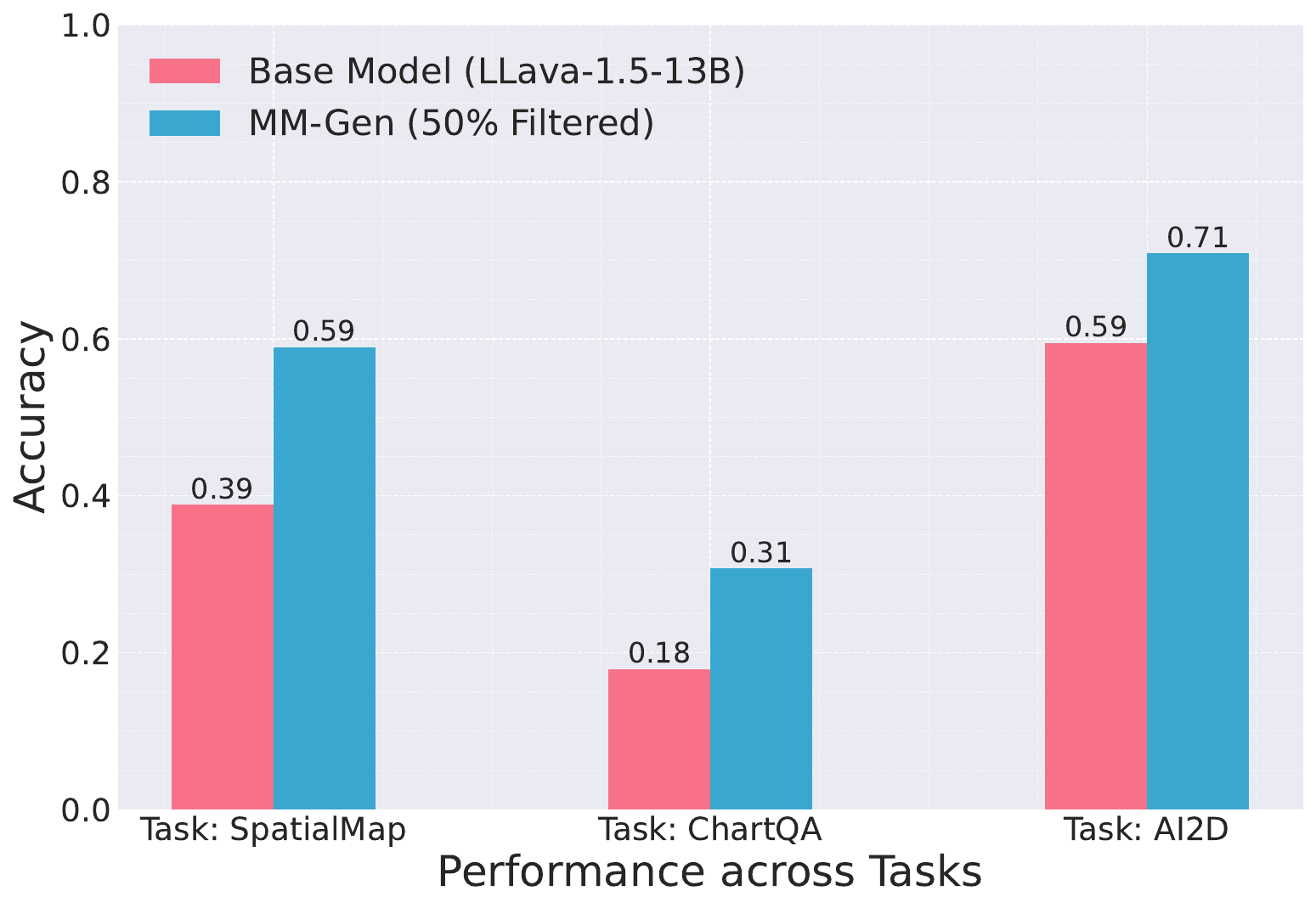}
    \caption{Evaluation on Llava-1.5 (13B Parameters)}
    \label{fig:13b}
\end{figure}

Fig.~\ref{fig:main_results} shows that \method{} can significantly improve upon the base model, across all 3 tasks, and either close the gap with or exceed the skyline performance across all three tasks. For ChartQA, \method{} achieves an absolute improvement of 15\% over the base model, reaching 0.5$\times$ of the skyline's improvement. For AI2D, \method{} shows a 14\% absolute improvement over the base model, and exceeds the skyline, achieving 1.6$\times$ the improvement that the skyline can obtain. Finally, on SpatialMap, \method{} demonstrates a 29\% absolute improvement over the base model which is 0.4$\times$ of the skyline's improvement. Figure \ref{fig:compare_baselines} shows a qualitative comparison of all baselines. 

Across all 3 tasks, we see that \method{} significantly outperforms baseline 2, highlighting the need for task-specific text annotations. Additionally, \method{} also outperforms the baseline 3: this emphasizes how it is crucial to specify the task in a data-centric manner i.e.,  using reference samples from the target task. It also highlights the significant challenge in specifying accurate and comprehensive task descriptions via natural language. Thus, not only is \method{} more easily generalizable across tasks, as it doesn't need significant human effort to describe the task, but it is also more effective. Interestingly, on ChartQA and SpatialMap, these baselines worsen the performance of the base model. 
Upon manual inspection of the generated data, we identified another limitation in the baselines: a lack of diversity (c.f. Appendix \ref{appendix:baseline_fail}) in the generated annotations. Consequently, training on such highly repetitive data can lead to model overfitting, diminishing the model's generalizability on these tasks. \method{} does not suffer from this as it creates diverse text annotations to mirror the diverse set of text annotations provided in the reference sample set. 

The spread of \method{}'s absolute improvements over baseline 1 can be attributed to the varying difficulties of each task for the base model, shown by the range of accuracies that the base model achieves on these tasks. Additionally, the varying improvements relative to the skyline can be explained by 1) the size and 2) the quality of the skyline data. For instance, on SpatialMap, the skyline performance is near-perfect, as the skyline data is created programmatically using the same code used to generate the test set and is thus perfectly i.i.d. In contrast, on AI2D and ChartQA, where data is curated by humans, the correspondence between training and test data is necessarily weaker. Moreover, the AI2D skyline dataset is relatively small ($\approx 5K$), which may contribute to its limited improvement. Despite these differences, \method{} consistently closes the gap to skyline performance, demonstrating for tasks in the wild, \method{} can curate task-relevant training data that is nearly as effective as human-curated data, with minimal human effort. Moreover, across all 3 tasks, we also observe that the 50\% filtered \method{} dataset nearly matches the performance of the larger, pre-filtered dataset while being twice as efficient for training. On AI2D, we even see a small improvement from filtering, likely due to reduced overfitting on redundant, unfiltered data. On SpatialMap, the relatively small drop (3\%) in filtered data performance can be attributed to the full dataset's higher diversity. This diversity arises from the nature of the task, where questions involving pairs or groups of objects scale combinatorially with the number of objects in the map, allowing for significant diversity in \method's generations.

Fig.~\ref{fig:13b} shows that, across all tasks, \method{} can even improve models as large as Llava-1.5 (13B Parameters). In fact, the resulting performance, across tasks, is even higher than that achieved by Llava-1.5 (7B parameters) in Fig. \ref{fig:main_results}. This shows that \method{} curated data can help boost performance of relatively stronger VLMs as well, utilizing their superior initial performance to achieve even higher performance, on target tasks. 

\begin{table}[!t]
    \centering
    \footnotesize
    \caption{Effect of Performance on Control Tasks (MMMU)}
    \vspace{-2mm}
    \label{tab:mmmu}
    \resizebox{0.75\linewidth}{!}{%
    \begin{tabular}{c|c} 
         \textbf{Model} & \textbf{Accuracy (\%)} \\ \midrule \midrule
         Base Model & 35.8 \\
         \method{} (ChartQA) & 33.6 \\ 
         \method{} (AI2D) & 37.0 \\ 
         \method{} (SpatialMap) & 34.1 \\ \bottomrule
    \end{tabular}
    }\vspace{-2mm}
\end{table}

\textbf{Performance on Control Tasks} In Table \ref{tab:mmmu}, we show that training on \method{} data, to improve performance on a given target task, does not hurt performance on other tasks (control tasks). Here, we use MMMU~\cite{yue2024mmmu} to represent these tasks as it considers a comprehensive evaluation of VLMs across many domains. 

\begin{table}[h]
    \centering
    \footnotesize
    \caption{Performance of Training on Combined \method Data.}
   \vspace{-2mm}
    \label{tab:combined}
    \resizebox{0.9\linewidth}{!}{%
    \begin{tabular}{c|c|c} 
         \textbf{Model} & \textbf{Base Model (\%)} & \textbf{\method{} All (\%)} \\ \midrule \midrule
         ChartQA & 18.2&  25.9\\
         AI2D & 55.2 & 65.7\\ 
         SpatialMap & 18.2 &  44.2\\ \bottomrule
    \end{tabular}
    }
\end{table}

\textbf{Combining Data from All Tasks} We also consider training Llava-1.5 (7B) in Table \ref{tab:combined} on a combination of data generated by \method for all tasks and observe that it can indeed improve performance across tasks simultaneously. 

\subsection{Ablations}

\begin{table}[t]
    \centering
    \caption{Ablation Study on \method{} using ChartQA}
    \vspace{-2mm}
    \label{tab:ablations_chartqa}
    \resizebox{0.95\linewidth}{!}{%
    \begin{tabular}{c|c} 
         \textbf{Ablation} & \textbf{Accuracy (\%)} \\ \midrule \midrule
         \method{} & 33.0 \\
         \method{} without Partition  & 31.6 \\ 
         \method{} with 3 In-Context Samples  & 30.5 \\ 
         10$\times$ Smaller Reference Set & 32.8 \\ \bottomrule
    \end{tabular}
    }
    \vspace{-5mm}
\end{table}

Here, we conduct ablations for \method{} on the chart understanding task (ChartQA). We vary different components of text annotation generation, and compare performance training on the resulting data. We do not filter the data here to isolate the differences in text generation.

\textbf{Importance of Partitioning into Subgroups}: Here, we investigate the importance of the partitioning into subgroups performed by \method{} prior to data generation by comparing performance with and without partitioning on ChartQA. As shown in Table \ref{tab:ablations_chartqa}, partitioning contributes a non-trivial 2\% of the total 15\% improvement that \method{} achieves.

\textbf{Effect of Number of In-Context Samples}: We assess the impact of varying the number of in-context samples provided to the stronger VLM during generation. As seen in Table \ref{tab:ablations_chartqa}, increasing the number of in-context samples from 1 to 3 actually decreases the final performance, likely due to the limitations of current VLMs on mutli-image understanding \cite{multiimage}. 

\textbf{Effect of Reference Sample Set Size}: Here, we compare the performance of \method{} using a 10$\times$ smaller reference sample set. Table \ref{tab:ablations_chartqa} shows that \method{} can still achieve nearly identical performance, highlighting how even a very small number of reference data is sufficient. 

\section{Conclusion}
In this paper, we introduced \method, a scalable and fully automated approach for curating task-specific multimodal data to enhance the performance of small vision-language models (VLMs) across specialized tasks. Our results demonstrate that \method achieves performance gains on specialized taksks, of up to 29\% absolute improvement over the base model and can even achieve 1.6x larger improvement compared to human-curated data (skyline), proving its efficacy in scenarios where human data curation is impractical. These results are a testimony of the untapped potential of multimodal data, and how automated and targeted text data enrichment can introduce improvements that cannot be harvested otherwise. Future avenues on this topic may study the value of creating planned curricula, when the goal is to create joint datasets that address task-specific gaps on a large number of tasks. In addition, in the absence of stronger teachers, it is also beneficial to study how approaches such as \method{} can use ensembles of several teachers in combination with answer verification techniques, to improve the quality of the training signal coming from synthesized data. 

\textbf{Acknowledgments} We sincerely thank Natasha Butt, Mazda Moayeri, Arindam Mitra, and Alessandro Stolfo for their valuable feedback and insightful discussions throughout this project. This research was partially supported by the National Science Foundation CAREER Award (Award No. 2146492), National Science Foundation Grant (Award No. 2421782), the Simons Foundation, Cisco Systems, Optum AI, the UCLA Hellman Fellowship, an Okawa Research Grant, and the Amazon Doctoral Fellowship.

\clearpage
{
    \small
    \bibliographystyle{ieeenat_fullname}
    \bibliography{references}
}

\onecolumn
\appendix
\newpage
\begin{center}
	\textbf{\LARGE Appendix }
\end{center}
\section{Exact Input to Stronger VLM and Generated Text Annotations}\label{app:examples}

\textbf{Exact Prompt to Stronger VLM}

\begin{lstlisting}
You are an expert in <name of task e.g. chart understanding / diagram understadning / spatial reasoning>. Given example image-question-answer tuples, 
your task is to generate diverse high-quality question-answer pairs relevant 
to this skill similar to the provided examples.

Step-by-Step Process:

1. Analyze the Example: Review the provided example question-answer pair to understand the structure, focus, and context.

2. Understand the New Image: Infer relevant details, objects, and themes in the new image, considering how they relate to the skill.
3. Generate Questions: Create questions that reflect the context and content of the new image, ensuring they align with the skill and follow the example's style.
4. If the question is a multiple-choice question, make sure to include the options in the question.
5. Formulate Answers: Generate accurate and concise answers to the questions. Ensure each answer directly corresponds to the content of the new image.

Output Format:
Return the results as a JSON list of objects. Each object should include:
- "Q": The generated question (include options if it's multiple-choice).
- "A": The generated answer.

Example Output:
[
  {"Q": "Generated question 1", "A": "Generated answer 1"},
  {"Q": "Generated question 2", "A": "Generated answer 2"}
]

<Refererence Sample> 

<Candidate Image>
\end{lstlisting}

\begin{figure}[!h]
    \centering
    \includegraphics[width=0.99\linewidth]{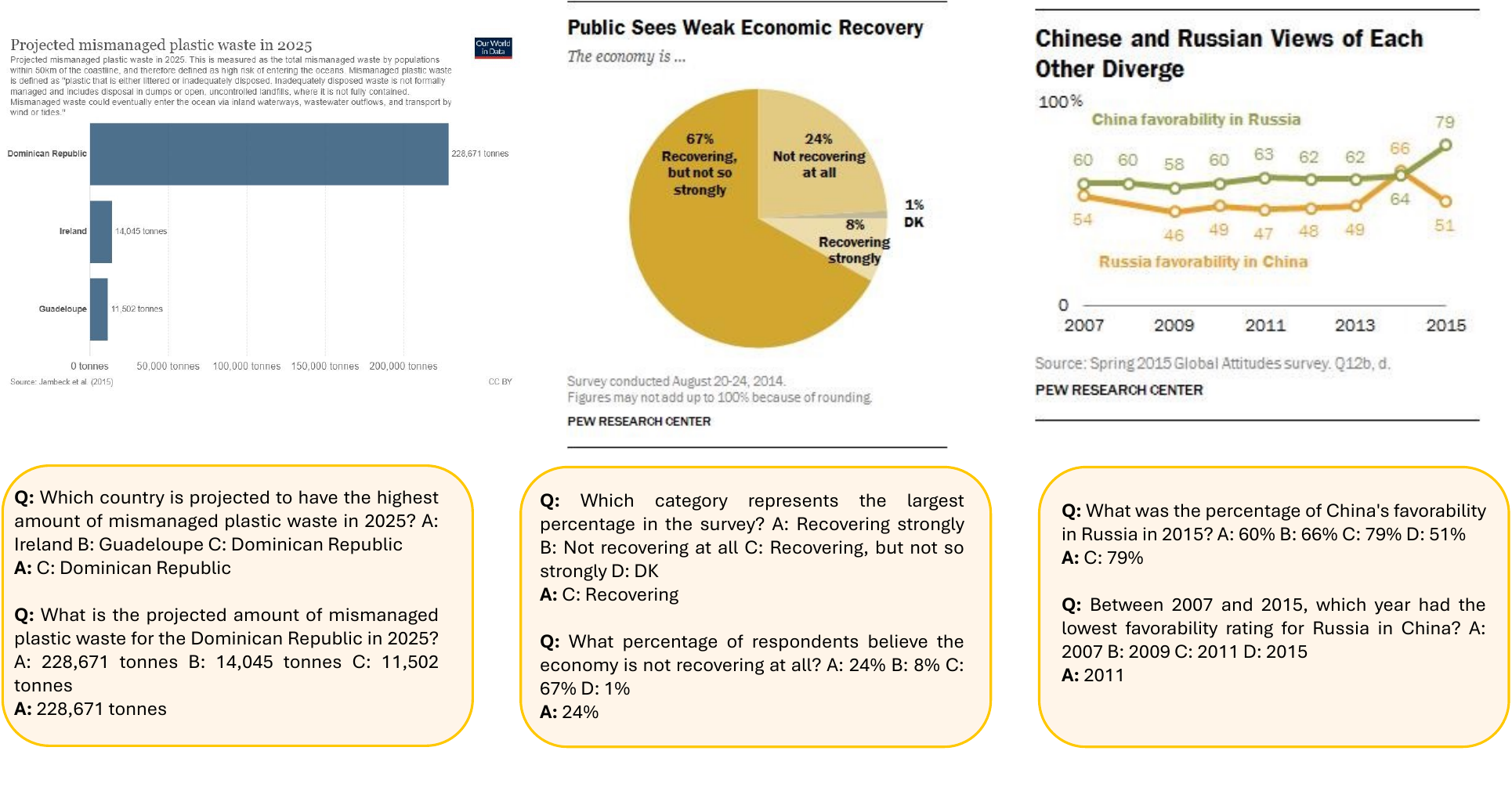}
    \caption{Examples Generated by \method for Chart Understanding}
    \label{fig:chartqa_examples}
\end{figure}

\begin{figure}[h]
    \centering
    \includegraphics[width=0.99\linewidth]{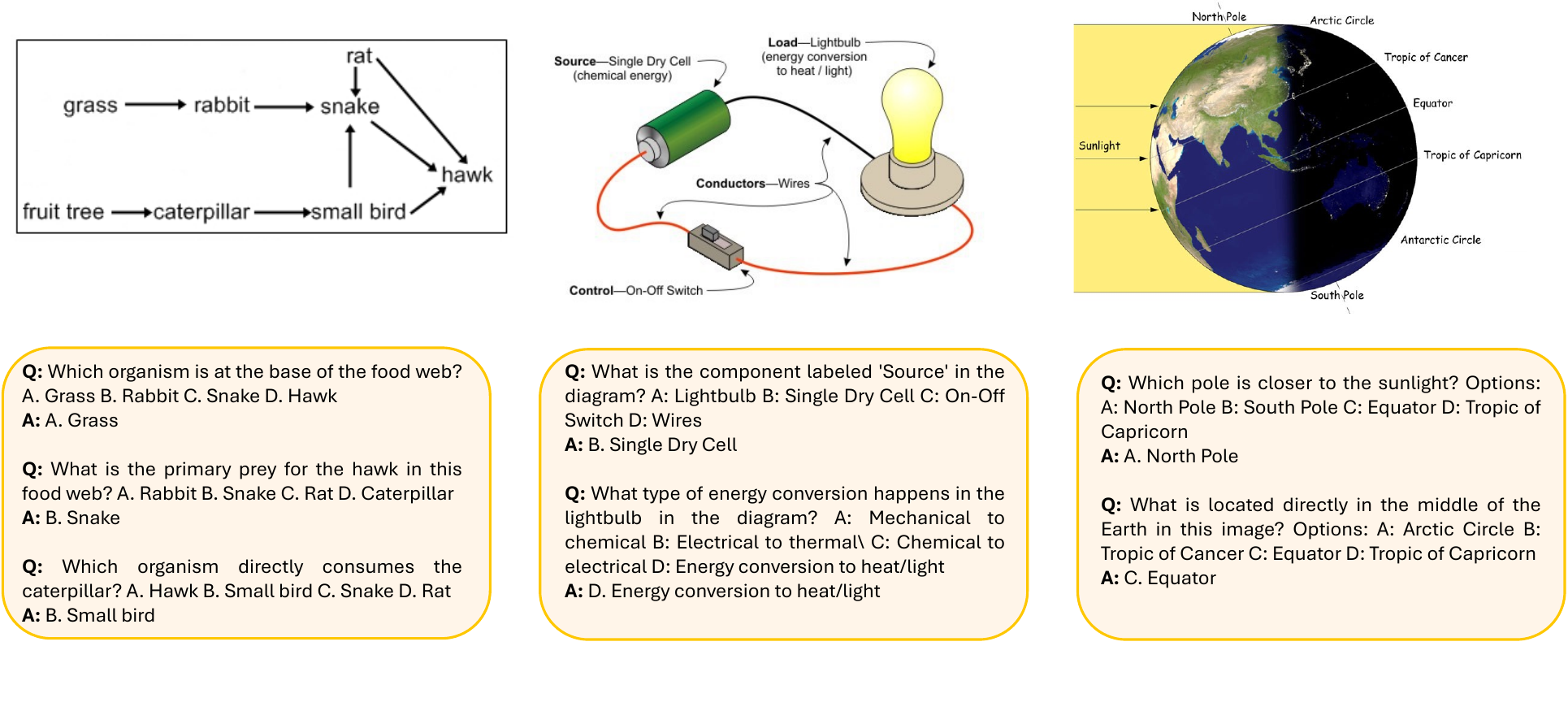}
    \caption{Examples Generated by \method for Diagram Understanding}
    \label{fig:ai2d_examples}
\end{figure}

\begin{figure}[t]
    \centering
    \includegraphics[width=0.99\linewidth]{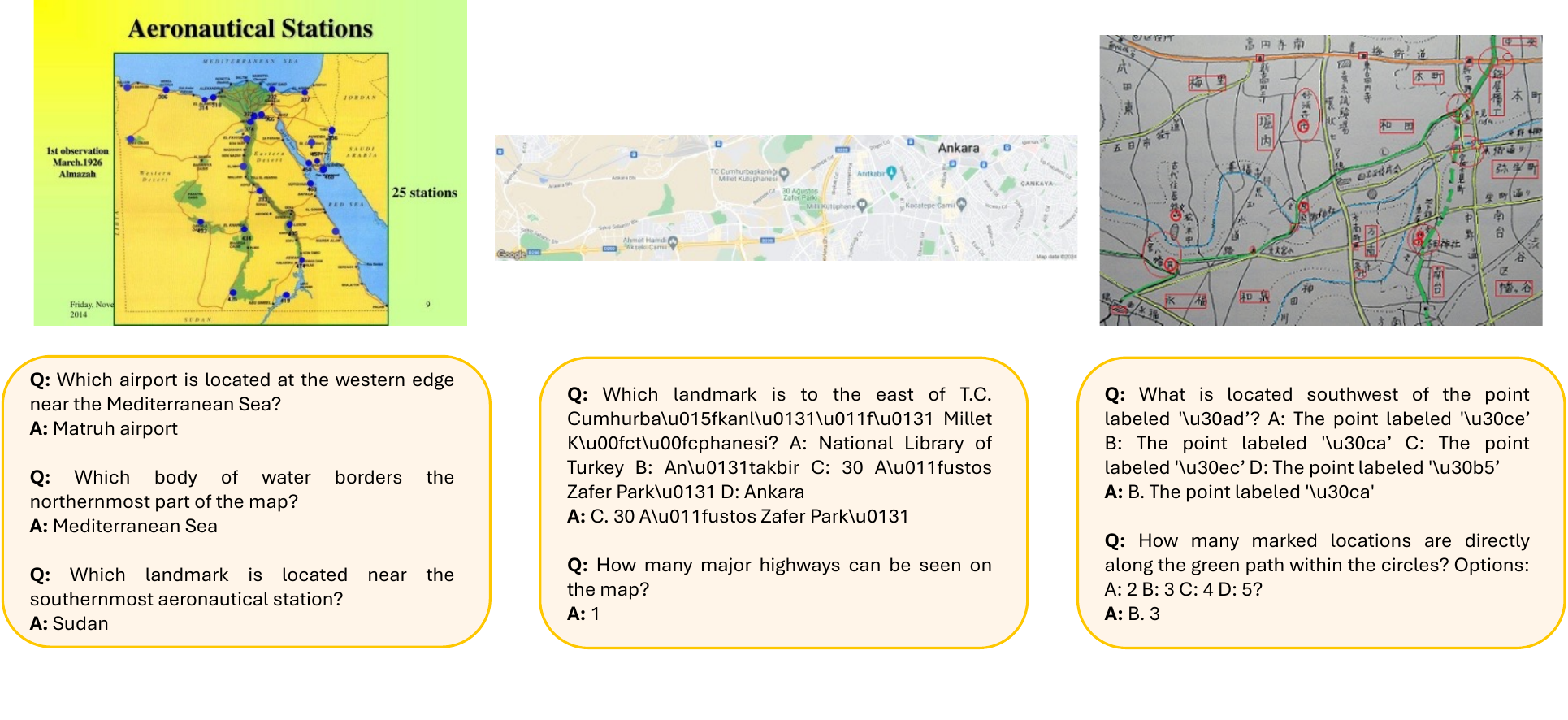}
    \caption{Examples Generated by \method for Spatial Reasoning on Maps}
    \label{fig:spatial_map_examples}
\end{figure}

Figure \ref{fig:chartqa_examples}, Figure \ref{fig:ai2d_examples} and Figure \ref{fig:spatial_map_examples} show examples generated by \method for chart understanding, diagram understanding and spatial reasoning on map, respectively.

\clearpage
\section{Pseudocode for \method}\label{app:alg}

In this section, we present the exact pseudocode for \method. Each of the three steps is denoted as a subroutine in the pseudocode. 

\begin{algorithm}
\caption{Data Generation Process}
\label{alg:data_generation}
\begin{algorithmic}[1]
\STATE \textbf{Subroutine 1: Partition} (\S~\ref{sec:method:partition})
\STATE $\{ (S^{\text{ref}}_{T_k}, V^{\text{pool}}_{T_k}) \}_{k \in \text{types}_T} = \textsc{PARTITION}(S^{\text{ref}}_T, V^{\text{pool}}_T, \text{types}_T)$ 

\STATE \textbf{Subroutine 2: Generate Data} (\S~\ref{sec:method:response_generation})
\FORALL{$k \in \text{types}_T$}
    \STATE $\mathcal{D}^{\text{GEN}}_k \leftarrow \emptyset$
    \STATE $\text{Iterator}(V^{\text{pool}}_{T_k}) \leftarrow$ Randomly order elements of $V^{\text{pool}}_{T_k}$ and create an infinite iterator
    \STATE Set $\text{NUM\_GEN\_PER\_REF} \leftarrow N \cdot \frac{|S^{\text{ref}}_{T_k}|}{|S^{\text{ref}}_T|}$
    
    \FORALL{$(v^{\text{ref}}, t^{\text{ref}}_p, t^{\text{ref}}_{\text{res}}) \in S^{\text{ref}}_{T_k}$}
        \FOR{$i = 1$ to $\text{NUM\_GEN\_PER\_REF}$}
            \STATE $v_{\text{candidate}} \leftarrow \text{NEXT}(\text{Iterator}(V^{\text{pool}}_{T_k}))$
            \STATE \footnotesize{$(t_p, t_{\text{res}}) \leftarrow L_{\text{VLM}}(\text{SYS\_PROMPT}, v^{\text{ref}}, t^{\text{ref}}_p, t^{\text{ref}}_{\text{res}}, v_{\text{candidate}})$}
            \STATE $\mathcal{D}^{\text{GEN}}_k \leftarrow \mathcal{D}^{\text{GEN}}_k \cup \{ (v_{\text{candidate}}, t_p, t_{\text{res}}) \}$
        \ENDFOR
    \ENDFOR
\ENDFOR
\STATE  $\mathcal{D}^{\text{GEN}} \leftarrow \bigcup_k \mathcal{D}^{\text{GEN}}_k$
\STATE \textbf{Subroutine 3: Filter} (\S~\ref{sec:method:filtering})
\STATE  $\mathcal{D}^{\text{GEN}_{\text{filt}}} \leftarrow$ Filter $\mathcal{D}^{\text{GEN}}$ by computing perplexity of all examples and selecting middle $r\%$ of examples
\RETURN $\mathcal{D}^{\text{GEN}_{\text{filt}}}$
\end{algorithmic}
\end{algorithm}

\clearpage
\section{Examples of Baselines Hurt Performance on Some Tasks} \label{appendix:baseline_fail}

\begin{figure}[h]
    \centering
    \includegraphics[width=0.99\linewidth]{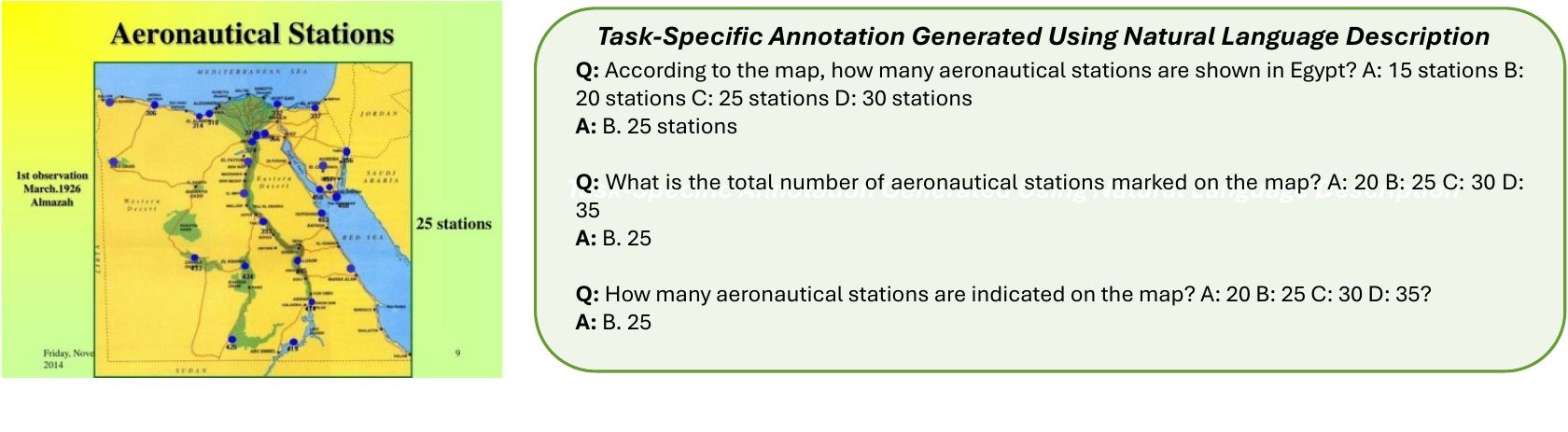}
    \caption{Task: Spatial Reasoning on Maps; \textbf{Highly Repetitive}  Generated Text for Baseline 3: Task-Specific Text Annotations, Generated Using Natural Language Task Description}
    \label{fig:spatial_map_1_fail}
\end{figure}

\begin{figure}[h]
    \centering
    \includegraphics[width=0.99\linewidth]{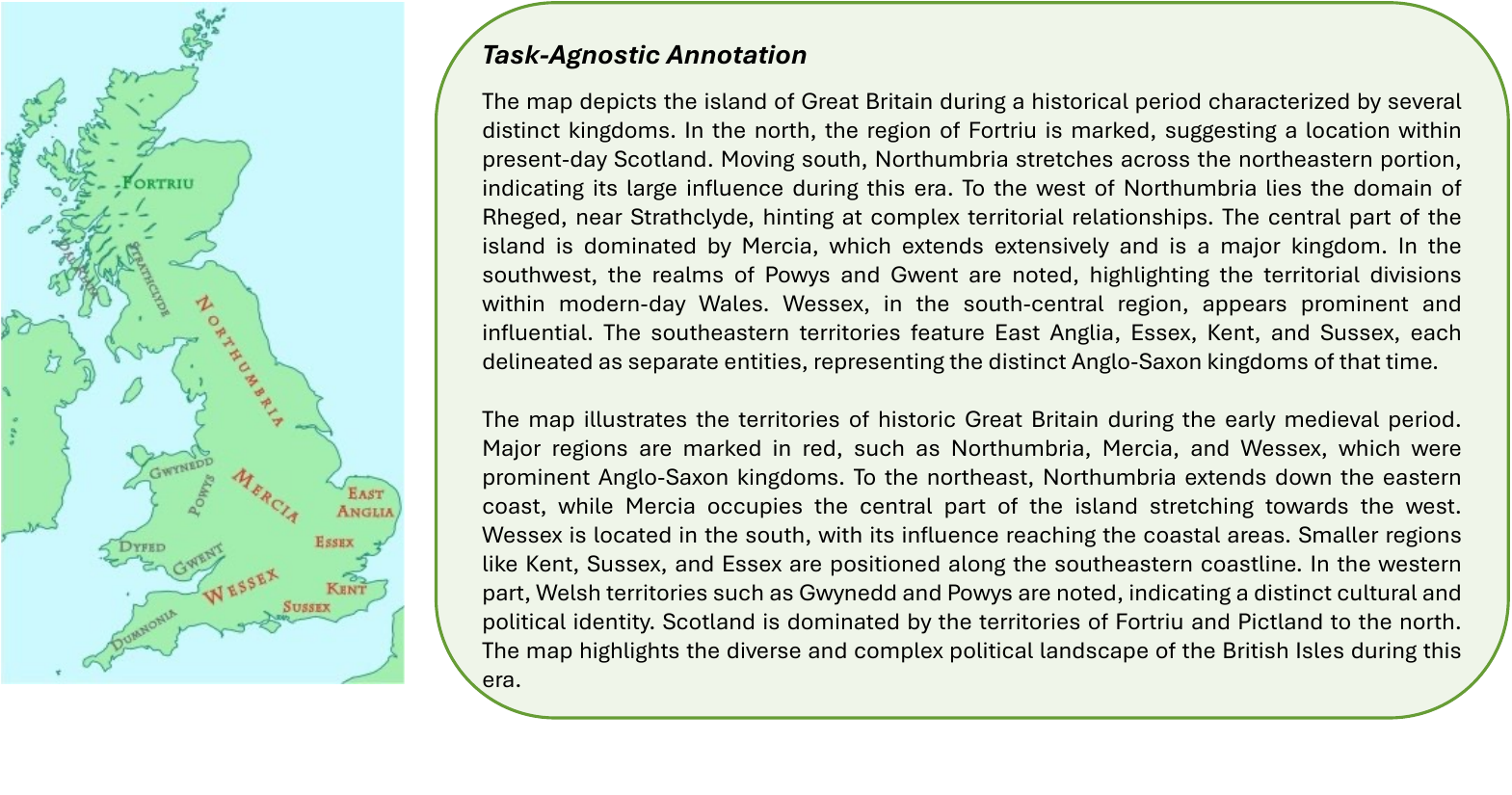}
    \caption{Task: Spatial Reasoning on Maps; \textbf{Highly Repetitive}  Generated Text for Baseline 2: Task-Agnostic Captions}\label{fig:spatial_map_2_fail}
\end{figure}

\begin{figure}[h]
    \centering
    \includegraphics[width=0.99\linewidth]{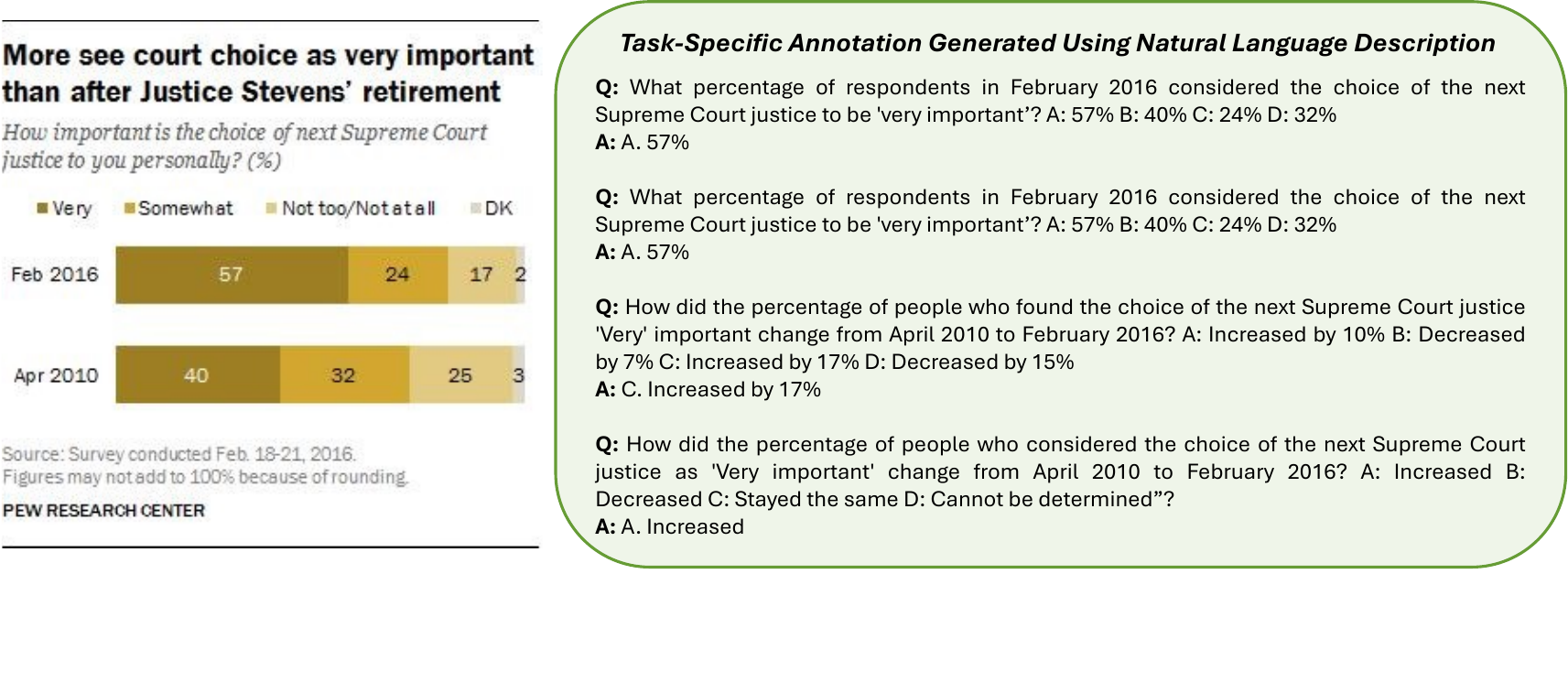}
    \caption{Task: Chart Understanding; \textbf{Highly Repetitive}  Generated Text for Baseline 3: Task-Specific Text Annotations, Generated Using Natural Language Task Description}
    \label{fig:chartqa_1_fail}
\end{figure}

In Sec. \ref{sec:experiments}, we observed that for ChartQA and for SpatialMap, some of the baselines can hurt the performance of the base model. Upon inspecting the generated data, we noticed that this is due to highly repetitive generated text annotations for these baselines. Here, in Figures \ref{fig:spatial_map_1_fail}, \ref{fig:spatial_map_2_fail}, \ref{fig:chartqa_1_fail}, we present some of the generated examples to highlight this problem.

\clearpage
\section{Data Generation and Training Details}\label{app:hparams}

For all data generation, we used the GPT-4o model \cite{gpt4} (2023-06-01-preview). 

For all the experiments, we use the follow common hyperparameters and trained on 4 A-100 GPUs. 

\begin{table}[ht]
    \centering
    \caption{Training Hyperparameters for \method{}}
    \label{tab:hparams}
    \begin{tabular}{l|c} 
         \textbf{Hyperparameter} & \textbf{Value} \\ \midrule \midrule
         Model Name or Path & liuhaotian/llava-v1.5-7b or liuhaotian/llava-v1.5-13b\\
         Vision Tower & openai/clip-vit-large-patch14-336 \\
         MM Projector Type & mlp2x\_gelu \\
         MM Vision Select Layer & -2 \\
         MM Use Image Start/End Token & False \\
         MM Use Image Patch Token & False \\
         Image Aspect Ratio & Pad \\
         Group by Modality Length & True \\
         BF16 & True \\
         Train Batch Size (Per Device) & 16 \\
         Eval Batch Size (Per Device) & 4 \\
         Gradient Accumulation Steps & 1 \\
         Learning Rate & 2e-5 \\
         Weight Decay & 0.0 \\
         Warmup Ratio & 0.03 \\
         LR Scheduler Type & Cosine \\
         TF32 & True \\
         Model Max Length & 2048 \\
    \end{tabular}
\end{table}
For each of the tasks, we tuned the number of epochs such that training loss converged for the \method generated data.  
\begin{enumerate}
\item Chart Understanding (ChartQA): 6 epochs
\item Diagram Understanding (AI2D): 6 epochs
\item Spatial Reasoning on Map (SpatialMap): 3 epochs
\end{enumerate}

\end{document}